%% file: main.tex
\documentclass[conference]{IEEEtran} 

\IEEEoverridecommandlockouts 

\usepackage{cite} 
\usepackage{amsmath,amssymb,amsfonts} 
\usepackage{algorithmic} 
\usepackage{graphicx} 
\usepackage{textcomp} 
\usepackage{xcolor} 
\usepackage{url} 
\usepackage{hyperref}
\usepackage{adjustbox}
\usepackage{cellspace}
\usepackage[final]{microtype}
\usepackage{mathptmx} 
\usepackage{comment}
\usepackage{pifont,xspace}

\newcommand{\one}{\ding{182}\xspace}
\newcommand{\two}{\ding{183}\xspace}
\newcommand{\three}{\ding{184}\xspace}
\newcommand{\four}{\ding{185}\xspace}
\newcommand{\five}{\ding{186}\xspace}

\def\BibTeX{{\rm B\kern-.05em{\sc i\kern-.025em b}\kern-.08em 
		
		T\kern-.1667em\lower.7ex\hbox{E}\kern-.125emX}} 

\begin{document} 
	
	\title{HybMT: Hybrid Meta-Predictor based ML Algorithm for Fast Test Vector Generation} 
	\author{\IEEEauthorblockN{Shruti Pandey} 	
		\IEEEauthorblockA{\textit{Department of Electrical Engineering} \\ 
			\textit{Indian Institute of Technology, Delhi}\\ 
			New Delhi, India \\ 
			shruti.pandey@ee.iitd.ac.in} 
		\and 
		\IEEEauthorblockN{Jayadeva} 		
		\IEEEauthorblockA{\textit{Department of Electrical Engineering} \\ 			
			\textit{Indian Institute of Technology, Delhi}\\ 			
			New Delhi, India\\ 			
			jayadeva@ee.iitd.ac.in} 		
		\and 		
		\IEEEauthorblockN{Smruti R. Sarangi} 		
		\IEEEauthorblockA{\textit{Department of Electrical Engineering} \\ 			
			\textit{Indian Institute of Technology, Delhi}\\ 			
			New Delhi, India \\ 			
			srsarangi@cse.iitd.ac.in} 		
	}
	\maketitle
	
	\input{abstract}

	\begin{IEEEkeywords}
		PODEM, stuck-at-fault, ATPG, ANN, ML
	\end{IEEEkeywords}
	
	\input{intro}
	\input{background}
	\input{methodology}

	\input{eval}

	\input{relatedwork}
	\input{conclusion}

	\bibliographystyle{IEEEtran}
	\bibliography{main}
	
\end{document}

%% file: abstract.tex
{\bf ML models are increasingly being used to
increase the test coverage and decrease the overall testing time.
This field is still in its nascent stage and up till now there were
no algorithms that could match or outperform commercial tools in terms of
speed and accuracy for large circuits.
We propose an ATPG algorithm HybMT in this paper that finally breaks
this barrier. Like sister methods, we augment the classical PODEM algorithm
that uses recursive backtracking. We design a custom 2-level predictor that
predicts the input net of a logic gate whose value needs to be set to ensure that the 
output is a given value (0 or 1). Our predictor chooses the output 
from among two first-level predictors, where the most effective one is a bespoke
neural network and the other is an SVM regressor.
As compared to a popular, state-of-the-art commercial ATPG tool, HybMT shows an overall
reduction of 56.6\% in the CPU time without compromising on the fault coverage for the EPFL benchmark circuits. HybMT
also shows a speedup of 126.4\% over the best ML-based algorithm while obtaining an equal or
better fault coverage for the EPFL benchmark circuits. }

%% file: intro.tex
\section{Introduction}
Testing VLSI circuits for stuck-at faults is a classical problem in computer
science. In its simplest form, the problem is equivalent to the SAT (circuit-satisfiability)
problem where we need to find the primary inputs (PIs) that set the value of a certain
net to 0 or 1. It is NP-Hard; hence, over the years a lot of heuristics have been developed
that ensure good test coverage and also try to minimize testing times as far as possible.
This area was considered to be thoroughly established and saturated at least a decade ago.

There are two reasons to look at this area in 2023. The first is that in the last 2-3 years,
a few encouraging results have been published that achieve good test coverage with ML-based
algorithms~\cite{roy2021training,roy2021principal,roy2021unsupervised}. The second is
that test-generation times have become very important with the ubiquity of ICs. 
For instance, Bloomberg reports~\cite{web:chipshortagebloomberg} that a delay of a few days can seriously
affect a product's commercial prospects quite
negatively, especially with regards to automotive chips.  

ML algorithms in this space are gradually maturing. They still have not moved beyond the 4-decade
old ISCAS '85 benchmarks and no comparisons with commercial tools have been made. 
We view our work HybMT as a logical successor in this line of work where we answer a simple
question, ``Is it possible to train a model on the 4-decade old ISCAS '85 circuits and 
generate test inputs for recently released benchmark circuits that are up to 70$\times$ larger 
and simultaneously outperform a popular
commercial ATPG tool?'' Note that a commercial tool has seen decades of research and
meeting its coverage metric or outperforming it 
in terms of execution time is very challenging.  We were successful and achieved a performance gain
of \textbf{56.6\%} without sacrificing test coverage. Even if we consider a counterfactual scenario where
our tool slightly underperformed vis-a-vis a proprietary tool, we argue that such research is still a
significant contribution because such algorithms enrich the open source ecosystem, and others can build on them.

Like other work in this area, we use an optimized version of the PODEM algorithm that is 
a part of the Fault~\cite{abdelatty2021fault} ATPG tool (part
of Google's OpenLane suite) as the base algorithm. An example of the key decision-making
process in PODEM is as follows: if an OR gate's output is `1', we 
need to decide which input needs to be set to `1'. We design a custom 2-level predictor (or meta-predictor)
that makes such predictions, which is known as {\em backtracing}. 
The second-level chooses the output of one of the 
first-level predictors: SVR regressor and a novel neural network that
we propose, \textit{HybNN}. 
~\footnote{The code and models are  available at this anonymous link:
\url{https://sites.google.com/view/vlsitesting22/home}}
	
Our specific contributions are as follows:
    \one Modifications to the PODEM algorithm to incorporate ML-based predictions for our {\em no-backtrack
probability}-based custom heuristic.
    \two Design and implementation of a novel 2-level predictor \textit{HybMT} based PODEM, which is highly accurate,
robust, and much faster than existing state-of-the-art approaches.
	\three Design of a novel meta-predictor that selects the best prediction model for each circuit net at runtime.
	\four Design of a novel neural network \textit{HybNN} to serve as a backtrace method in PODEM.
    \five Demonstration of significant speedup over a commercial ATPG tool (56.6\%) 
and the existing state-of-the-art ML-based algorithm (126.4\%).

Our paper is organized as follows: Section~\ref{sec:back} presents the background of the PODEM algorithm and other key
concepts involved in our work. We present the methodology in Section~\ref{sec:meth} followed by 
experiments and results in Section~\ref{sec:eval}. Finally, we present the related work in Section~\ref{sec:rel} and
conclude in Section~\ref{sec:conc}.

%% file: background.tex
\section{Background}
\label{sec:back}
\subsection{Path Oriented Decision Making (PODEM)} 

Path Oriented Decision Making (PODEM) \cite{goel1981implicit} is a
fault-oriented deterministic test generation (DTG) algorithm that works by making assignments only at the PIs.
PODEM finds the PI values to activate the fault and propagate it to the \textit{primary output} (PO).
To do this, PODEM traverses the circuit till the PI (\textit{backtracing}) and reverses a previous decision if this
decision conflicts with a newer objective (\textit{backtracking}).


\subsection{Design for Testability Measures}
\textit{Controllability/Observability Program (COP):} The COP  controllability\cite{roy2020machine} is defined as
the probability of setting a net $l$ to logic 1 or 0 by making random PI assignments (each input net has a 50\% probability of
being 1). {\em Observability} is defined as the probability of propagating the value at a net to a primary output (PO).


 \textit{Sandia Controllability/Observability Analysis Program (SCOAP):}
The SCOAP \cite{roy2020machine} combinational controllability is defined as the difficulty of setting a
net's logic level to 1/0. It is represented as CC1/CC0 (resp.). It is the minimum number of PIs required to set the
value of net $x$ to a given value  $v$.  The SCOAP combinational observability is defined as the difficulty of observing
the value at an internal net at a PO. It is denoted as CO. It is the minimum number of PI assignments that need to be
made to propagate the fault to a PO.  Note that SCOAP measures do not take reconvergent fanouts into account. 


%% file: methodology.tex
\section{ Methodology } \label{sec:meth}

\subsection{Implementation Details}
\begin{figure}[ht!]
  \centering
  \includegraphics[width=0.5\textwidth]{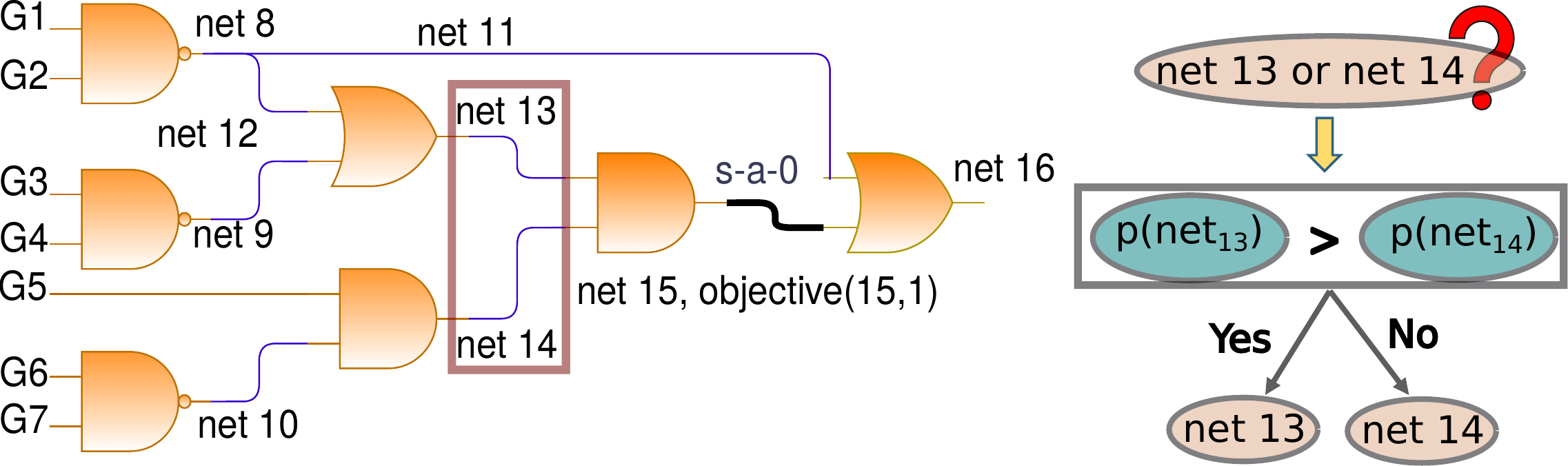}
  \caption{Decision making during backtracing and the use of `probability of no-backtrack' ($p$) as a heuristic}
  \label{fig:metav1}
\end{figure}

Figure~\ref{fig:metav1} illustrates the decision-making involved during backtracing and the use of
\textit{no-backtrack probability} (\textit{p}) as a heuristic. Our proposed ATPG algorithm selects the net with a higher
value of $p$ (as estimated by HybMT) to avoid wasted work.


\subsubsection{\textbf{Input Features of the ML Model}}
We use the following input features to train our predictors.

\emph{\textbf{The COP Combinational Controllability of a Net $l$ ($CC(l)$):}}
First, we assign the CC values for all the PIs to 0.5 as the PIs need to take the values ‘1’
and ‘0’ randomly. We use the truth tables of different gates to derive the formulae for computing the CC values for
the respective output nets of these gates. Then, we calculate the CC values for the intermediate nets and POs of the
circuit by traversing the circuit graph in the forward direction. 

\textit{\textbf{The COP Combinational Observability of a Net $l$ ($CO(l))$:}}
First, we assign the CO values for all the POs to ‘1’ as any error at the POs is always
observable. The formulae for computing the CO values for the input nets of different gates, given the CO values of the
output net and CC values of the other input nets, are derived using the truth tables of these gates. Then, we calculate
the CO values for the intermediate nets and PIs of the circuit by traversing the circuit graph in the backward direction
and using the derived formulae.

\textit{\textbf{The shortest distance of a net from the PIs:}}
The distance feature is normalized to bring the values in the range $[0,1]$ using the following formula 
\begin{equation}
    distance(l)' = \frac{distance(l) - distance_{min}}{distance_{max} - distance_{min}}
\end{equation}

where $distance_{min}$ and $distance_{max}$ are the minimum and maximum values for the distance feature computed across all the circuit net samples, respectively.

\textit{\textbf{The type of the gate:}}
The supported gate types are PI, PO, PPI, PPO, NOT, AND, NAND, OR, NOR, XOR, XNOR, DFF, BUF and BAD. The
type-of-the-gate feature is represented using 14 bits; we use one-hot encoding. PPI and PPO stand for pseudo-primary
inputs and pseudo-primary outputs, respectively, while DFF and BUF stand for D flip-flops and buffers. The gate type BAD
represents gates that do not belong to any of the specified gate types.

\textit{\textbf{SCOAP zero controllability (CC0) and SCOAP one controllabillity (CC1):}} First, we assign the CC0 and CC1 values
for all the PIs to 1 and set the depth of each of the logic gates to 1. We compute the sum of the SCOAP controllability
values of different input net assignments that justify the desired value at the output net. Then, we take the minimum of
these sums and add it to the depth of the gate to get the CC0/CC1 value at the output of the respective logic gate.
Finally, we use the CC0 and CC1 values of the logic gate outputs to figure out the CC0 and CC1 values of all the nets by
going through the circuit graph in the forward direction. 

\textit{\textbf{SCOAP observability (SCOAP CO):}} First, we assign the CO values for all the POs to 0. Then, we
compute the CO value of a logic gate's input as the sum of the CC0/CC1 values of the other input nets, the CO value of
the gate's output net, and the depth of the gate. The lower the values of the SCOAP measures for a net, the easier it is
to control and observe that net.

\textit{\textbf{Fanout of the gate:}} The fanout of a gate is the number of logic gate inputs driven by the output of
the logic gate under consideration.  

The input features are shown in Table \ref{tab:features}. Our lower-level predictors (ML models and HybNN) require the first four features whereas HybMT requires all the eight features.
\begin{table}[h]
	\caption{Input Features of the Predictors}
	\label{tab:features}
	\centering
	\resizebox{\columnwidth}{!}{
		\begin{tabular}{|c|c|c|}
			\hline
			 COP controllability & COP observability & 	Distance from the PI \\ \hline
		 The type of the gate & SCOAP CCO & SCOAP CC1 \\ \hline
		SCOAP observability & Fanout of the gate & \\ \hline
			
	\end{tabular}}
	
\end{table}

\subsubsection{\textbf{Procedure for Finding Top-$k$ Hard-to-detect Faults in a Circuit}}
The detection probability ($P_{detect}$) of a fault can be found using Equations~\eqref{eq:sa0} and
\eqref{eq:sa1}. Insight: The detection probability of an s-a-v (stuck at $v$) fault takes into account
the probability of setting a value of $\overline{v}$ at the faulty net (fault activation) and the probability of propagating the fault to a
PO (given by $CO(l)$).
 
\begin{equation} \label{eq:sa0}
    P_{detect} = CC(l) \times CO(l) \text{ for s-a-0 fault at net\textit{ l} }
\end{equation}
\begin{equation} \label{eq:sa1}
    P_{detect} = (1-CC(l)) \times CO(l) \text{ for s-a-1 fault at net \textit{l} }
\end{equation}

We sort the fault list in an ascending order of the detection probability and thus estimate the top $k$ (hard to
detect) faults. 

\subsubsection{\textbf{Output of the Model}} 
We generated the ground truth label for each circuit net by running PODEM with COP controllability values as a heuristic
while backtracing for $k$ hard-to-detect faults of the circuit. During the backtracing procedure of PODEM, if a PI
assignment does not lead to any conflict, we label the nets involved in the backtrace as ‘1’, else as ‘0’. A
circuit net occurs several times during a single run of PODEM. The input features of a circuit net remain the same
across its occurrences, but the ground truth label may keep changing. Hence, we computed the probability of no backtrack
$(p)$ for a circuit net $l$ as follows:

\begin{equation}
p = \frac{f(label_l = 1)}{f_l}
\end{equation}

Here, $f(label_l = 1)$ is the frequency of the occurrence of a `1' in the label for the net $l$ and $f_l$ is the
frequency with which the net $l$ appears in the data. The attributes of the model's output are mentioned in Table
\ref{table:output}.

\begin{table}[h!]
\caption{The output of the lower-level predictor (ML model) and its attributes}
\label{table:output}
\centering
\resizebox{0.5\textwidth}{!}{
\begin{tabular}{@{}|c |c| c| c| @{}}
\hline
{\textbf{ Output Features}} & { \textbf{Data Type}} & { \textbf{Size (in bytes)}} & { \textbf{Range}} \\ \hline
Probability of no-backtrack            & float                            & 4                                      & [0, 1]                       \\ \hline
\end{tabular}}
\end{table}

\subsection{Network Architectures}

\begin{figure}[h!]
	\centering
	\includegraphics[width=0.5\textwidth]{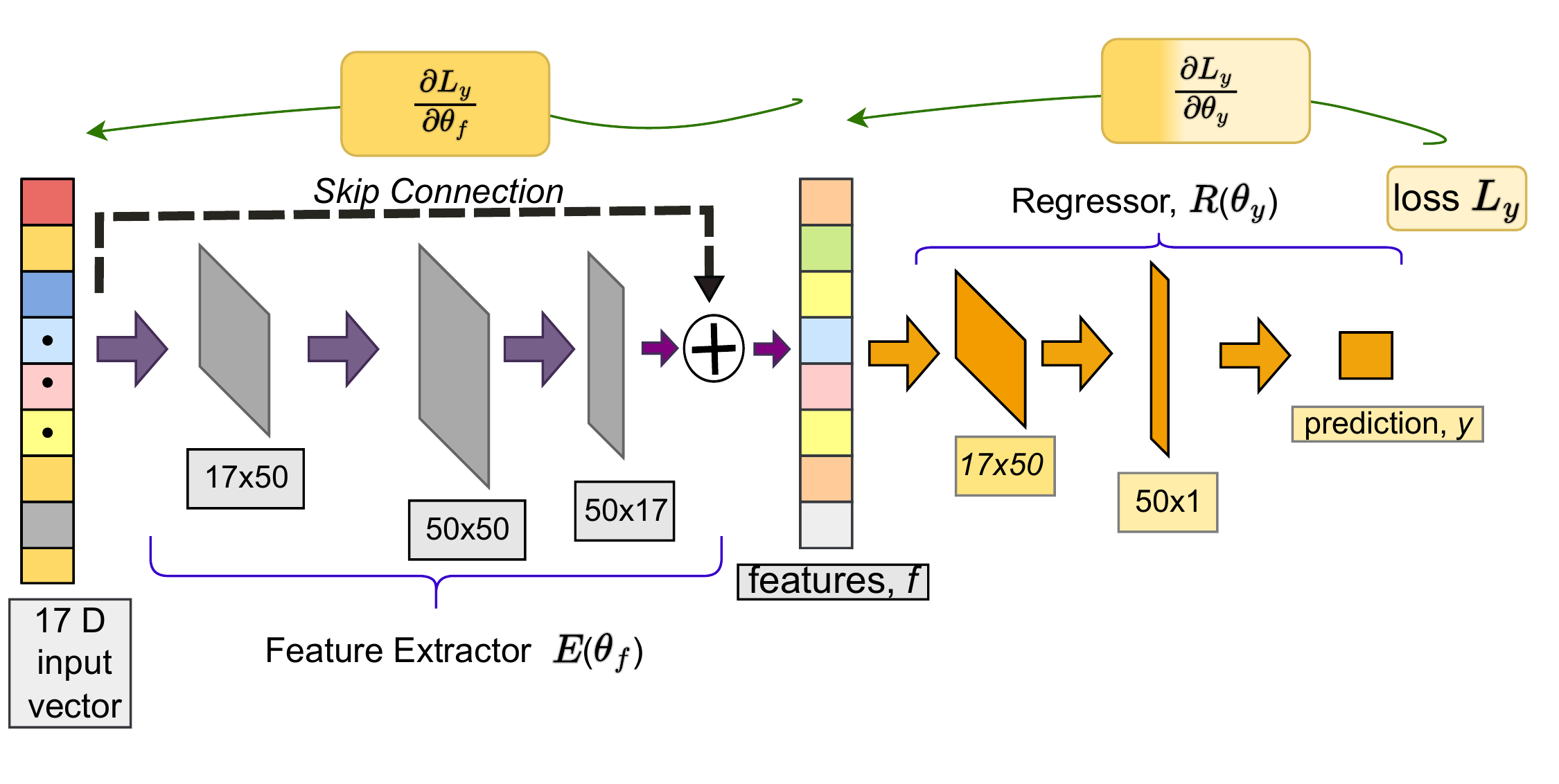}
	\caption{The architecture of the proposed HybNN model with its sub-networks: feature extractor and regressor. }
	\label{fig:hybnn}
\end{figure}

\textbf{Hybrid Neural Network (\textit{HybNN}):}
We designed a novel fully connected feed-forward neural network with a skip connection, which we call \textit{HybNN}. It has two sub-networks:
a feature extractor $E(\theta_f)$ and a regressor {$R(\theta_f)$}. Here, $\theta_f$ and $\theta_y$ represent the
learnable parameters of the respective sub-networks. The feature extractor takes the features of a circuit net as the
inputs and performs further extraction to learn the most significant features. We define a linear layer followed by a
nonlinear activation function, \textit{rectified linear unit (ReLU)}, as a hidden layer. Every nonlinear layer adds to
the complexity and the learning capability of the network. Hence, deeper networks with more hidden layers are found to
perform well \cite{montufar2014number}. $E(\theta_f)$ has two such hidden layers. We also incorporated a skip connection ~\cite{he2016deep} that adds the input features directly to the output of $E(\theta_f)$. The resulting sum is then fed to the function $R(\theta_y)$. $R(\theta_y)$ consists of a single hidden layer and an output layer
followed by a \textit{sigmoid} layer. The output of the regressor network is the no-backtrack probability, which is used
as the backtracing heuristic. The detailed architecture of \textit{HybNN} is shown in Fig. \ref{fig:hybnn}. The change
in the loss function with respect to the model parameters is given by $\partial{L_y} / \partial\theta_y$ and
$\partial{L_y}/\partial\theta_f$ for the regressor and the feature extractor, respectively. We propagate the gradients backwards as shown in Fig. ~\ref{fig:hybnn} to train HybNN. With the increase in the number of layers, the gradient becomes very small or zero near the initial layers, and thus the initial layers do not get updated. This is termed as the vanishing gradient problem. A skip connection is used to solve the vanishing gradient problem as it provides an alternate path for the gradient to flow ~\cite{he2016deep}. We see a significant improvement in performance by adding a skip connection to our neural network. 

\begin{figure}[t!] 

\centering 

\includegraphics[width=0.5\textwidth]{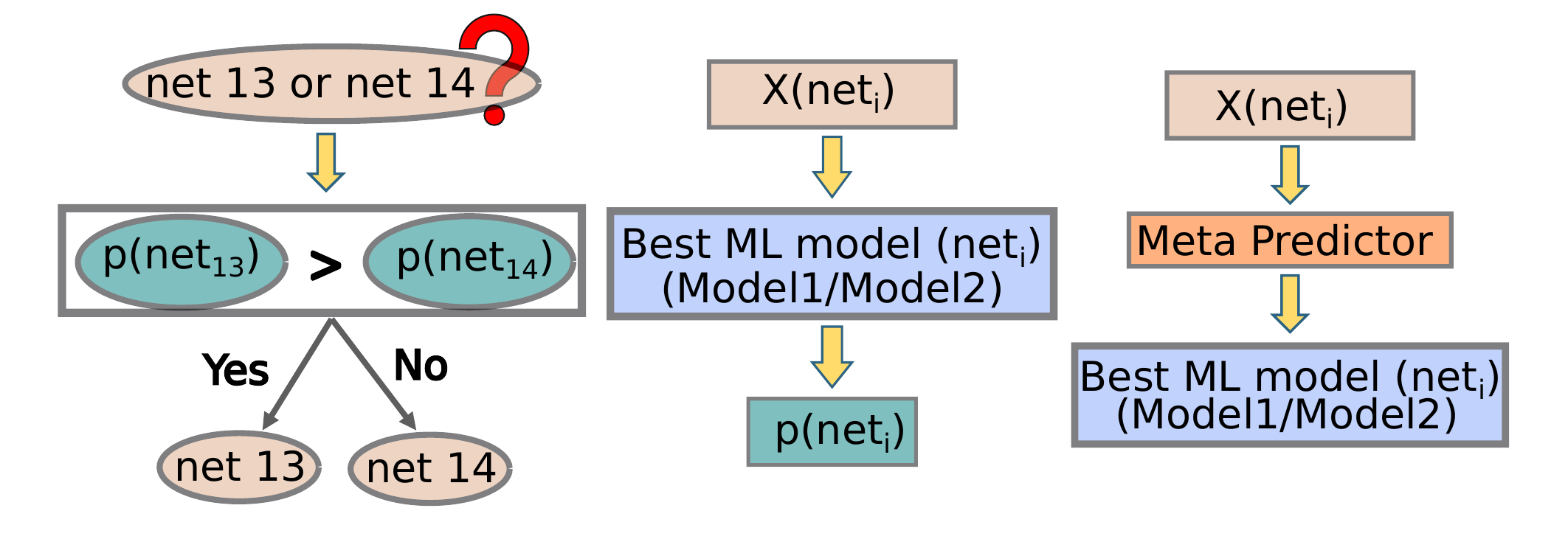} 

\caption{The concept of the HybMT model. $X(net_i)$ represents the input features of the $i^{th}$ net.} 

\label{fig:metav2} 

\end{figure}

\textbf{Hybrid Meta Predictor (\textit{HybMT}):}
A meta-predictor model learns to choose among the best predictor models for a given test instance at runtime. We
designed a hybrid meta-predictor \textit{HybMT} that selects between the two best performing
lower-level models. The inputs to the \textit{HybMT} model are the features of a circuit net described in
Table~\ref{tab:features} and the output is a class label corresponding to the best-performing model. \textit{HybMT} is a
\textit{random forest classifier} (RFC) model designed for a binary classification task. An RFC model is based on an
ensemble learning technique. It combines the predictions of different decision tree classifiers to give a more
accurate prediction than its constituent models.  
We get the no-backtrack probability $(p)$ for each net from the lower-level model corresponding to the
predicted class. Next, we use $p$ to guide the backtracing step in PODEM. The concept of \textit{HybMT}-based PODEM for
the circuit of Fig. \ref{fig:metav1} is illustrated in Fig. \ref{fig:metav2}.

%% file: eval.tex
\section{Experiments and Results}
\label{sec:eval}


\subsection{Basic Setup}
We modified an open-source ATPG framework for PODEM obtained from Fault \cite{abdelatty2021fault}, a part of Google's
OpenLane project \cite{ghazy2020openlane} to incorporate our changes.  We used Flex v2.6.4 and Bison v3.0.4 for parsing
the net lists in the ISCAS \cite{brglez1985neutral} format. We implemented the ANNs using PyTorch v1.10.2.  We conducted
experiments to find the optimal learning rate of the Adam optimizer and set it to 0.01.  The baseline is a faithful
implementation of the ML algorithm described in reference~\cite{roy2021training}. We have used a {\em standard setup}
for evaluating our scheme~\cite{roy2021training}. The same setup including the same set of benchmarks have been used in
recent papers in the area.


\textbf{Data Preparation:} Training and test data points must be entirely different (and disjoint) to ensure the
correctness and generalizability of the ML models. Hence, we follow a scheme of leave-one-out testing where we keep one
benchmark circuit separate for performing inference and train on the remaining benchmark circuits (for every benchmark). 
We always train our data on the arcane ISCAS 85~\cite{brglez1985neutral} benchmark suite, and test on both the ISCAS benchmarks and
the much larger and recent EPFL circuits~\cite{amaru2015epfl}.

\textbf{Cross-Validation:} We further divide the training set obtained from the leave-one-out scheme into $k$ equal
parts. We keep one of the parts as the validation set, and the other parts constitute the training set. We evaluate the
models on the validation set. We repeat this procedure for each data fold. Then, the average accuracy or loss is
computed across the $k$-folds. As each of the data points is used for validating the model, the cross-validation score
shows the model's generalization capability. This is the standard practice.

\subsection{Lower-level Predictors: Experiments and Results}
\subsubsection{Traditional ML Models}

We started with performing experiments with different regression models: regularized linear regressors (LASSO and
Ridge), Support Vector Regressor (SVR), Decision Tree Regressor (DTR) and Random Forest Regressor (RFR) for the
no-backtrack probability prediction task. We used 5-fold cross-validation to tune the hyperparameters of these models.
The models, their hyperparameters, and the choice of the range of values for tuning are shown in Table \ref{tab:hp}
(for the ISCAS suite).

\begin{table}[!htb]
	\caption{Models for regression with their hyperparameters and ranges for tuning}
	\label{tab:hp}
	\centering
	\resizebox{0.95\columnwidth}{!}{
		\begin{tabular}{|c|c|c|}
			\hline
			\textbf{Models} & \textbf{Hypeparameters}  &  \textbf{Range} 
			\\\hline
			Lasso & Regularization coefficient ($\alpha$) & 100 values in the range [0.1,1]  \\ \hline
			Ridge & Regularization coefficient ($\alpha$) & 100 values in the range [0.1,1]  \\ \hline
			SVR &  Regularization coefficient ($C$)  & [1e-3, 1e4]   \\ \hline
			DTR & Depth of the tree  & [1,31] \\ \hline
			RFR & Number of decision trees  &  [1,101] in steps of 10  \\ \hline
			
	\end{tabular}}
\end{table}

\subsubsection{Our Proposed \textbf{Hy}brid \textbf{N}eural \textbf{N}etwork (HybNN)}
The loss function and the optimizer used for training the \textit{HybNN} model are \textit{mean squared error} (MSE) and
\textit{Adam}, respectively. We  used a uniform distribution in the range $(-\sqrt{k}, \sqrt{k})$ to initialize the
model weights, where $k = \frac{1}{\text{\#input features}}$. 

\subsubsection{Execution Times for the Lower-Level Predictors}
\begin{figure}[htp!] 
	\centering 
	\includegraphics[width=0.5\textwidth]{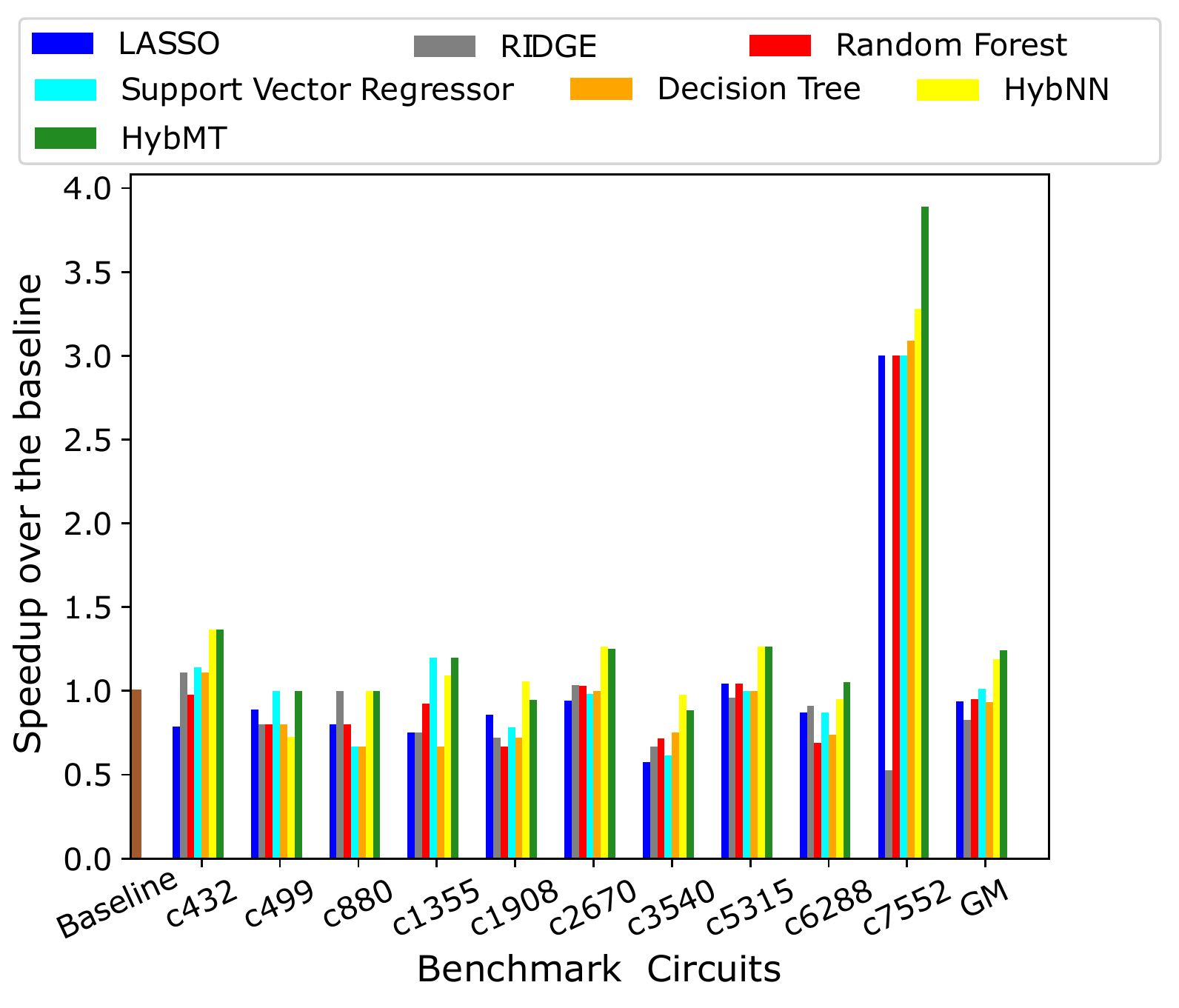} 
	\caption{Speedup obtained over the baseline for 100 hard-to-detect faults (ISCAS'85 suite)} 
	\label{fig:resreg} 
\end{figure}

We ran PODEM on 100 hard-to-detect faults of the ISCAS circuits (results shown in Fig.~\ref{fig:resreg}). Let us
ignore the results for HybMT (last bar in each cluster) for the time being. We observe that the two best-performing
models in terms of their highest geometric mean speedups are HybNN and SVR, respectively.  \textit{HybNN} shows a
reduction of \textbf{20.4\%} in CPU time as compared to the baseline, and SVR shows a mean reduction of 1.2\%.  
Out of the 10 benchmarks, HybNN is the best performing model in 8/10 and for the remaining two,
SVR is the best.

\textbf{Motivation for the HybMT model:} 
These results indicate that there is no unanimous winner among all the ML models that we considered even though HybNN
comes close. A hybrid meta-predictor is thus a promising idea as long as it correctly chooses between HybNN and SVR.

\subsection{Results with a Commercial ATPG Tool and EPFL Benchmarks}
We ran experiments on a commercial ATPG tool for 100 random faults of the EPFL benchmark circuits~\cite{amaru2015epfl}
(arithmetic circuits, up to 70$\times$ larger than ISCAS), and compared them with the baseline and HybMT (results shown in Table~\ref{table:epflres}).
Unlike the baseline and HybMT, the commercial tool does not have the notion of hard-to-detect faults; we thus
choose 100 random faults (clearly putting ourselves at a disadvantage).
Note that we consider the same fault list for both HybMT and
baseline.  Our HybMT algorithm obtains a speedup of \textbf{56.6\%} over the commercial tool while obtaining the same
(or similar) fault
coverage.  We also performed experiments with the baseline~\cite{roy2021training} on the EPFL circuits and observed that HybMT has a 
speed-up of \textbf{126.4\%} over the baseline. In terms of coverage, we are roughly neck-and-neck with the commercial
tool. In
general, we are better than the commercial tool by roughly 2\% (fault coverage).
Table \ref{table:epflbacktracks} shows a comparison between the number of backtraces and backtracks obtained with HybMT and the baseline for the EPFL circuits. Note that the commercial tool does not report these metrics. We observed that the sum of backtraces and backtracks is proportional to the CPU time. Table \ref{table:epflbacktracks} shows that HybMT has a much smaller number of backtraces and backtracks than the baseline for most circuits. This justifies the significant speedup obtained by HybMT.

\begin{figure}[!htb]
	\centering
	\includegraphics[width=0.5\textwidth]{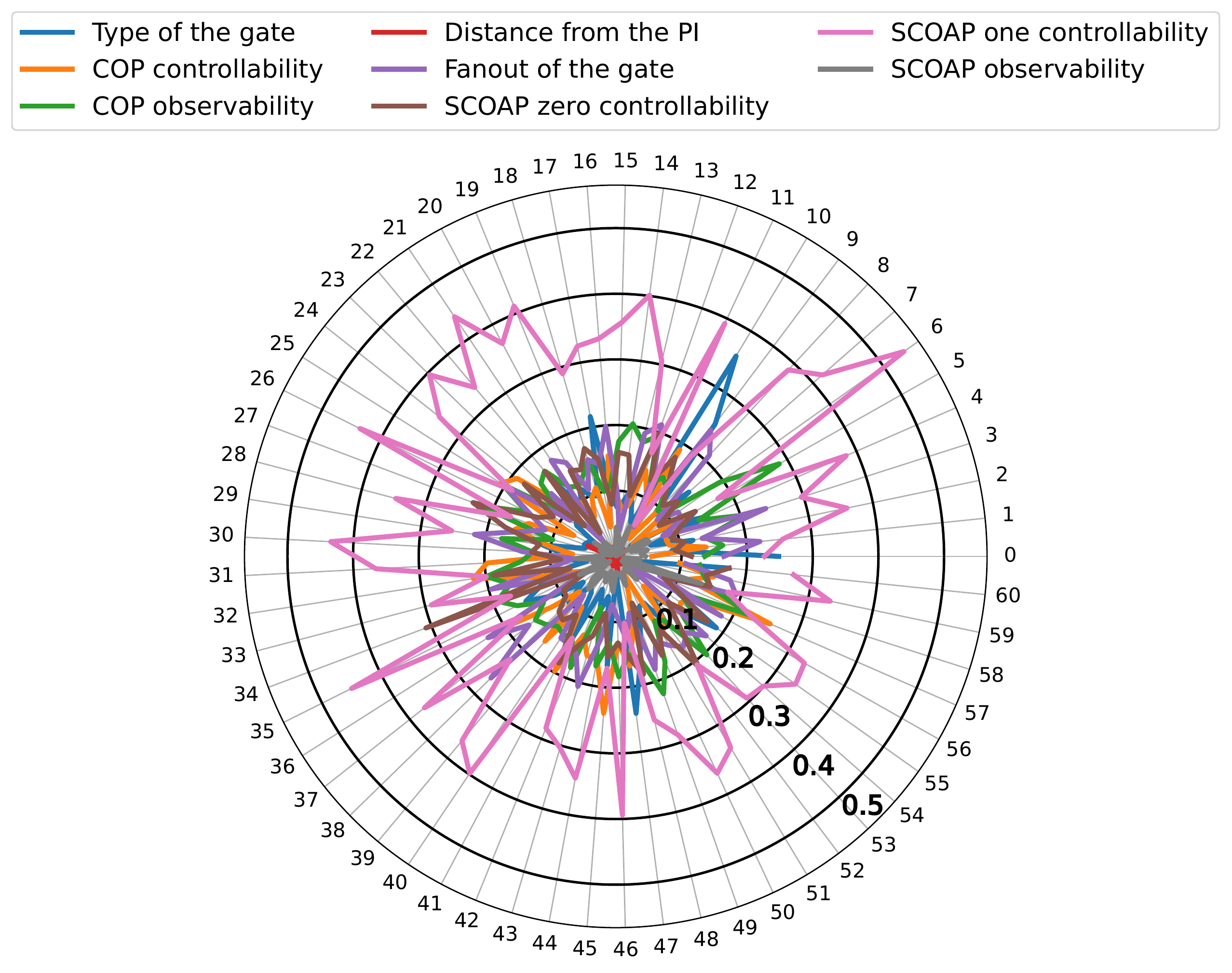}
	\caption{Importance of the features in the decision trees of
		the RFC meta predictor model}
	\label{fig:radar}
\end{figure}

The radar plot of Fig.~\ref{fig:radar} shows the importance of each feature in the decision making process of the
decision trees of the random forest model (for lower-level predictor selection using HybMT). The greater the importance
of a feature, the more it is used in making decisions for the data samples. The $angular$-axis represents the decision trees of
the random forest part of the \textit{HybMT} model, and the $radial$-axis represents the number of times a feature is used to split a decision tree node,
weighted by the number of samples it splits. We observed that {\em SCOAP one controllability} is the most important
feature of all. The results are trying to say that the character of a net is primarily determined by the minimum number
of PIs that can control it. Next in importance are the type of the gate and COP observability.

\begin{table*}[!htb]
	\caption{Comparison with a commercial tool for the EPFL arithmetic benchmark circuits}
	\label{table:epflres}
	\begin{center}
		
		
		\resizebox{0.75\textwidth}{!}{
			\begin{tabular}{|*{7}{c|}}
				\hline
				& \multicolumn{3}{c}{\textbf{CPU Time}} & \multicolumn{3}{|c|}{\textbf{Fault Coverage}}\\
				\textbf{Benchmark} & \multicolumn{3}{c}{\textbf{(in ms)}} &\multicolumn{3}{|c|}{\textbf{(in \%)}} \\
				\cline{2-7}
				\textbf{Circuits} & \textbf{Commercial ATPG Tool} & \textbf{Baseline} & \textit{\textbf{HybMT}} & \textbf{Commercial ATPG Tool} & \textbf{Baseline} & \textit{\textbf{HybMT}} \\
				\hline
				\hline
				Adder & 190 & 2,891 &\textbf{180} & 98 & 100 &  \textbf{100}\\
				\hline
				Barrel-shifter & 340 & 50 & \textbf{55} & 98 & 100 & \textbf{100} \\
				\hline
				Divisor & 46,480 & 46,756 & \textbf{46,455} & 98 & 97 & \textbf{98} \\
				\hline
				Max & 950 & 44 &\textbf{780} & 98 & 100 & \textbf{100}  \\
				\hline
				Multiplier & 2,510 & 2,545 & \textbf{2,200} & 98 & 100 & \textbf{100} \\
				\hline
				Sine & 790 & 3,25,340 & \textbf{708} & 97 & 74 &\textbf{ 97 }\\
				\hline
				Square & 1,870 & 647 & \textbf{800}& 98 & 98 & \textbf{98} \\
				\hline
				
		\end{tabular} }
	\end{center}
\end{table*}

\begin{table}[!htb]
	\caption{Comparison with the baseline for the EPFL arithmetic benchmark circuits}
	\label{table:epflbacktracks}
		
		
		\resizebox{0.5\textwidth}{!}{
			\begin{tabular}{|*{5}{c|}}
				\hline
				\textbf{Benchmark}	& \multicolumn{2}{c}{\textbf{Baseline}} & \multicolumn{2}{|c|}{\textit{\textbf{HybMT}}}\\
				\cline{2-5}
				\textbf{Circuits} & \textbf{\#Backtraces} & \textbf{\#Backtracks} & \textbf{\#Backtraces} & \textbf{\#Backtracks}\\
				\hline
				\hline
				Adder & 1,19,809	& 0	& 15,723 &	0 \\
				\hline
				Barrel-shifter & 1,064 &	0 & 1,078 &	0 \\
				\hline
				Divisor & 1,191,655 &	4,84,780 &	1,165,223 &	3,23,653 \\			
				\hline
				Max & 1,546	& 0 &	2,869 &	0 \\			
				\hline
				Multiplier & 27,319 &	239 &	21,832 &	105 \\			
				\hline
				Sine & 11,347,743 &	5,653,684 &	7,853 &	352 \\
				\hline
				Square & 11,880 &	0 &	10,850 & 	0 \\			
				\hline
				
		\end{tabular} }
\end{table}
\subsection{HybMT: Training and Basic Inference Results (ISCAS '85)}
We generated the ground truth labels for the meta predictor (\textit{HybMT}) model in two ways. In the first approach,
we record the number of times a net is backtracked while running PODEM using different ML model-based probabilities as a
heuristic. We take the model that gives the minimum number of backtracks for a net as its class label.  Recall that by
analyzing the CPU time improvements of the lower-level regressors, we found that \textit{HybNN} and SVR performed the
best for 8 and 2 benchmark circuits, respectively. Hence, in the second approach, we assigned a class label of 0 and 1
to all the nets of the circuits where \textit{HybNN} and SVR perform the best (resp.).

We obtained better results with the second approach -- a coarse-grained approach. The results also corroborate the fact
that generating the ground truth labels at the circuit level captures a global view of the circuit's structure and hence
gives better results.  The \textit{HybMT} meta-predictor (top level) achieves an average 5-fold cross-validation
accuracy of\textbf{ 99\%} for the low-level predictor selection task. The results show that our proposed \textit{HybMT} model
reduces the CPU time of PODEM by \textbf{26.8\%} over the baseline for the ISCAS'85 benchmark circuits.

%% file: relatedwork.tex
\section{Related Work} \label{sec:rel} 
Our primary focus was on recent work (published in the last three years) in the area of ML-based
ATPG algorithms that target stuck-at faults. Almost all the highly cited and well regarded work 
in this field follows the same line of thinking, which is to reduce the number of backtracks in highly optimized versions of PODEM.

 In 2020, Roy et al. \cite{roy2020machine} used the output of ANNs as a backtracing heuristic in PODEM. This was the seminal paper (in recent times). A year later, Roy et al. \cite{roy2021training} proposed improvements in the training strategy. In this approach, the backtrace history of a fault is only included in the training data if it decreases the number of backtracks in the ANN-guided PODEM algorithm and the training data is generated by running PODEM on both easy and hard to detect faults of the circuits.

In these works, the ANN used the following input features: distance from the PI (primary input), testability measures and the type of the gate driven by the line. The experiments were performed on 100 hard-to-detect faults of the ISCAS'85 \cite{brglez1985neutral} and ITC'99 \cite{roy2020machine} benchmark circuits. 
Subsequently, Roy et al.~\cite{roy2021unsupervised} proposed a method based on unsupervised learning to combine different heuristics using \textit{principal component} (PCA) analysis. The major PCs are used for guiding the backtracing in PODEM. This method also uses the testability measures from SCOAP \cite{roy2020machine} and the fanout count as the input features. PCA-assisted supervised learning has also been explored in \cite{roy2021principal} to reduce the complexity of the ANN used for guiding the backtracing in PODEM. So far, no work has presented a robust data generation technique or demonstrated consistent improvement across all benchmark circuits. Thus, they highlight a need for a more robust, accurate and better approach.

In this paper, we \underline{compare} our work with the latest work in this series that experimentally provided the best results, which is reference~\cite{roy2021training}.  




%% file: conclusion.tex
\section{Conclusion}\label{sec:conc}
The aim of this paper was to reduce the CPU time of the PODEM algorithm as much as possible without sacrificing the fault coverage for the most {\em hard-to-detect} faults. Our novel two-level predictor \textit{HybMT} model based PODEM obtains a speedup of \textbf{56.6\%} over a commercial tool for the EPFL benchmarks. HybMT also outperforms the existing state-of-the-art algorithm \cite{roy2021training} by \textbf{26.8\%} and \textbf{126.4\%} for the ISCAS'85 and EPFL benchmarks, respectively. We used the lower-level models to predict the \textit{no-backtrack probability} $(p)$, and then we used $p$ to guide backtracing in PODEM. We experimented with different traditional ML and neural network models to serve as the lower level models. We found that our novel \textit{HybNN} model and the \textit{support vector regressor} (SVR) models perform the best for different benchmark circuits. Our work shows a significant reduction in the CPU time of PODEM while achieving \textbf{high fault coverage} for the ISCAS'85 \cite{brglez1985neutral} and EPFL \cite{amaru2015epfl} benchmark circuits.

We further conclude that in our work, a global circuit-level view is required to learn better backtracing heuristics. Properly designed neural networks are capable of learning better features and hence are a better candidate for guiding backtracing. The explanations for the meta predictor model indicate that the SCOAP one controllability occurs the maximum number of times in the decision making process of the decision trees of the random forest classifier and hence is the most important feature. Now, our immediate aim is to increase the expressive power of the model by raising the limit of 70$\times$ (test size/train size) to 100$\times$, and look
at even larger circuits with sequential elements.